\documentclass{article}
\usepackage{spconf,amsmath,graphicx}
\usepackage{amsfonts}
\usepackage{multirow}
\usepackage{cite}
\usepackage{svg}
\usepackage{url}
\usepackage{xcolor}

\usepackage{hyperref}

\title{TOWARDS ACCURATE CROSS-DOMAIN IN-BED HUMAN POSE ESTIMATION}
\name{\begin{tabular}{c}Mohamed Afham$^{\star \dagger}$,  Udith Haputhanthri$^{\star \dagger}$,  Jathurshan Pradeepkumar$^{\star \dagger}$,  Mithunjha Anandakumar$^{\dagger}$ \\
Ashwin De Silva$^{\ddagger \dagger}$,  Chamira U. S. Edussooriya$^{\dagger \mathsection}$ \thanks{$^{\star}$ These authors contributed equally to the work. \newline $\copyright$ 2021 IEEE. Personal use of this material is permitted. Permission from IEEE must be obtained for all other uses, in any current or future media, including reprinting/republishing this material for advertising or promotional purposes, creating new collective works, for resale or redistribution to servers or lists, or reuse of any copyrighted component of this work in other works.}\end{tabular}}
  
\address{$^{\dagger}$ Dept. of Electronic and Telecommunication Engineering, Univeristy of Moratuwa, Sri Lanka \\ $^{\ddagger}$Dept. of Biomedical Engineering, Johns Hopkins University, Baltimore, MD, USA \\ $^{\mathsection}$ Dept. of Electrical and Computer Engineering, Florida International University, Miami, FL, USA }

\begin{document}
%
\maketitle
\begin{abstract}
Human behavioral monitoring during sleep is essential for various medical applications. Majority of the contactless human pose estimation algorithms are based on RGB modality, causing ineffectiveness in in-bed pose estimation due to occlusions by blankets and varying illumination conditions. Long-wavelength infrared (LWIR) modality based pose estimation algorithms overcome the aforementioned challenges; however, ground truth pose generations by a human annotator under such conditions are not feasible. A feasible solution to address this issue is to transfer the knowledge learned from images with pose labels and  no occlusions, and adapt it towards real world conditions (occlusions due to blankets). In this paper, we propose a novel learning strategy comprises of two-fold data augmentation to reduce the cross-domain discrepancy and knowledge distillation to learn the distribution of unlabeled images in real world conditions. Our experiments and analysis show the effectiveness of our approach over multiple standard human pose estimation baselines. Our code is available at: \url{https://github.com/MohamedAfham/CD_HPE}.

\end{abstract}


\vspace{-0.5em}
\section{Introduction}
\label{sec:intro}

An individual spends roughly a third of their lifetime at rest in the bed. Human behavioral monitoring during sleep is crucial for prognostic, diagnostic, and treatment of many healthcare complications, where in-bed postures are the fundamental factor. As an example, the in-bed postures of patients who have undergone surgeries should be monitored continuously long-term, where maintaining correct postures will lead to quick recovery \cite{seeingunder}.  Visual inspections by the caretaker \cite{Pressureulcers} are widely used for in-bed pose estimations, but this is labor intensive, and the interpretations will be subjective. Wearable devices have been utilized for in-bed pose estimation, but their obtrusive nature towards subjects degrades sleep quality. 

Recent advancements in computer vision have enabled contactless camera-based human pose estimation. Camera-based methods are less expensive, comfortable for the subject, and require less maintenance. Since the introduction of convolutional pose machine \cite{CNNposemachine}, there have been many works related to 2D human pose estimations \cite{simplebaseline, stackHG,sunetal,Yangetal} and 3D human pose estimations \cite{3d}, where the algorithms have achieved high performance. Some works have addressed human pose estimation under general settings such as multi-person pose estimation \cite{multisurvey} and wild setting \cite{wild}. However, the majority of the studies related to pose estimation are based on RGB imaging modality. Considering the in-bed pose estimation setting, RGB imaging modality-based algorithms becomes ineffective due to: 1) illumination conditions, 2) presence of heavy occlusions due to the blankets (covered subjects) and 3) privacy of the subjects, which could cause problems during large scale data collections. As a result, only a few researches have focused on in-bed pose estimation \cite{seeingunder,liu2020}. 

The thermal diffusion-based long-wavelength infrared (LWIR) imaging approach for in-bed human behavioral monitoring proposed in \cite{seeingunder,liu2020} overcomes the aforementioned drawbacks in RGB imaging modality-based in-bed pose estimation. However, obtaining annotations (ground truth poses) for the covered images is infeasible in real-world applications as the occlusions and illumination conditions may vary. The realistic way to tackle this challenge is to develop a robust in-bed human pose estimation algorithm by only utilizing labeled LWIR images of uncovered (without blankets) subjects and unlabeled LWIR images of covered (with blankets) subjects during training. The trained algorithm should be able to achieve accurate pose estimation on real-world covered LWIR images.

This challenge can be addressed by utilizing unsupervised domain adaptation techniques in in-bed human pose estimation algorithms. An inherent drawback of deep neural networks is their lack of ability to learn representations from out-of-distribution data \cite{domainshift}. Domain adaptation overcomes this challenge by transferring the knowledge learned in the source domain to the target domain. Several works \cite{DASurvey} have been proposed in the classification paradigm to learn the representations of out-of-distribution data. Particularly \cite{uda_kd1, uda_kd2} utilize a knowledge distillation strategy in image classification to learn out-of-distribution data. However, to the best of our knowledge only \cite{regda} has proposed a domain adaptation strategy in a regression setting to improve keypoint detection. In contrast, our novel data augmentation along with knowledge distillation-based domain adaptation in the in-bed pose estimation setting is the \emph{first} of such works. In this paper, we propose a novel learning strategy to adapt existing human pose estimation algorithms to achieve our end goal. Our key contributions are as follows: 1) two-fold data augmentation strategy to reduce discrepancies between different domains, 2) self-supervised knowledge distillation for cross-domain in-bed pose estimation and finally 3) component wise quantitative performance analysis.

\vspace{-2ex}
\section{Methodology}
\label{sec:method}

\begin{figure}[t!]
    \vspace{-1.5em}
    \centering
    \includegraphics[width = 0.9 \linewidth]{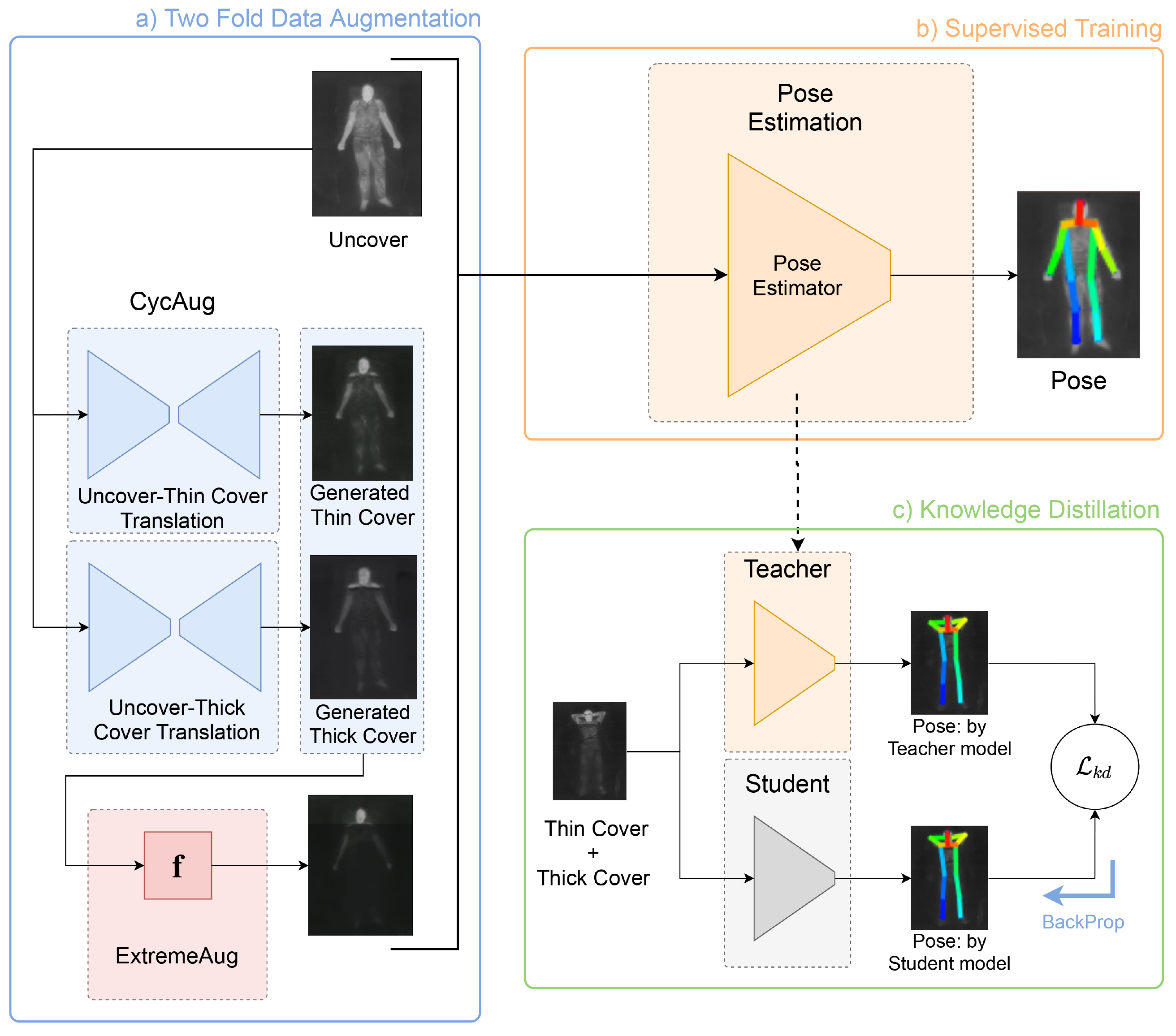}
    \caption{Overall architecture of the proposed method. We augment the uncovered images using the two-fold data augmentation strategy to reduce the cross domain discrepancy. Knowledge distillation in a self-supervised manner further improves the performance by learning transferable features.}
    \label{fig:overall}
    \vspace{-1.32em}
\end{figure}

We formulate the problem statement in Sec. \ref{sec:problem}; then we introduce our specific data augmentation protocol in Sec. \ref{sec:data_aug}, and finally we present our pose estimation technique and knowledge distillation in Sec. \ref{sec:pose_estimation} and \ref{sec:kd} respectively.

\vspace{-0.5em}
\subsection{Problem Definition}   \label{sec:problem}
In this paper, we aim to tackle the problem of cross-domain in-bed pose estimation. In standard 2D human pose estimation tasks, we have access to a dataset $\mathcal{D}$ with $\mathnormal{n}$ labeled samples $\{x_i, y_i\}_{i=1}^{n}$, where  $(x_i, y_i) \in \mathcal{X} \times \mathcal{Y}_K$. Here, $\mathcal{X}$ is the input space and $\mathcal{Y}_K$ $\in$ $\mathbb{R}^{2 \times K}$ is the output space and \textit{K} is the number of keypoints under consideration. The goal is to learn a function $\textit{f}_\theta$ which minimizes the error rate $\textit{E}_\mathcal{D}=\mathbb{E}_{(x,y\sim\mathcal{D})}\mathcal{L}(f_{\theta}(x),y))$, where $\mathcal{L}$ is the loss function. In a cross-domain setting, we have access to a labeled source domain dataset $\mathcal{D}_s = \{x_{i}^s, y_{i}^s\}_{i=1}^{n^s}$ and an unlabeled target domain dataset $\mathcal{D}_t = \{x_{i}^t\}_{i=1}^{n^t}$. The objective of standard unsupervised domain adaptation, is to minimize $\textit{E}_{\mathcal{D}_t}$ through the representations learnt from $\mathcal{D}_s$. In the context of this work, we consider the labeled uncovered LWIR imaging modality data as $\mathcal{D}_s$ and the unlabeled covered LWIR imaging modality data as $\mathcal{D}_t$. To demonstrate the generalizability of our proposed method under challenging conditions, we evaluate our method on the following sub-types of the target datasets: $\mathcal{D}_{t_1}$ which consists of thin covered LWIR images (a thin sheet with approximately $1$ mm thickness) and $\mathcal{D}_{t_2}$ which consists of thick covered LWIR images (a thick blanket with approximately $3$ mm thickness). 


\begin{figure}[t!]
    \vspace{-1.2em}
    \centering
    \includegraphics[width = 0.92 \linewidth]{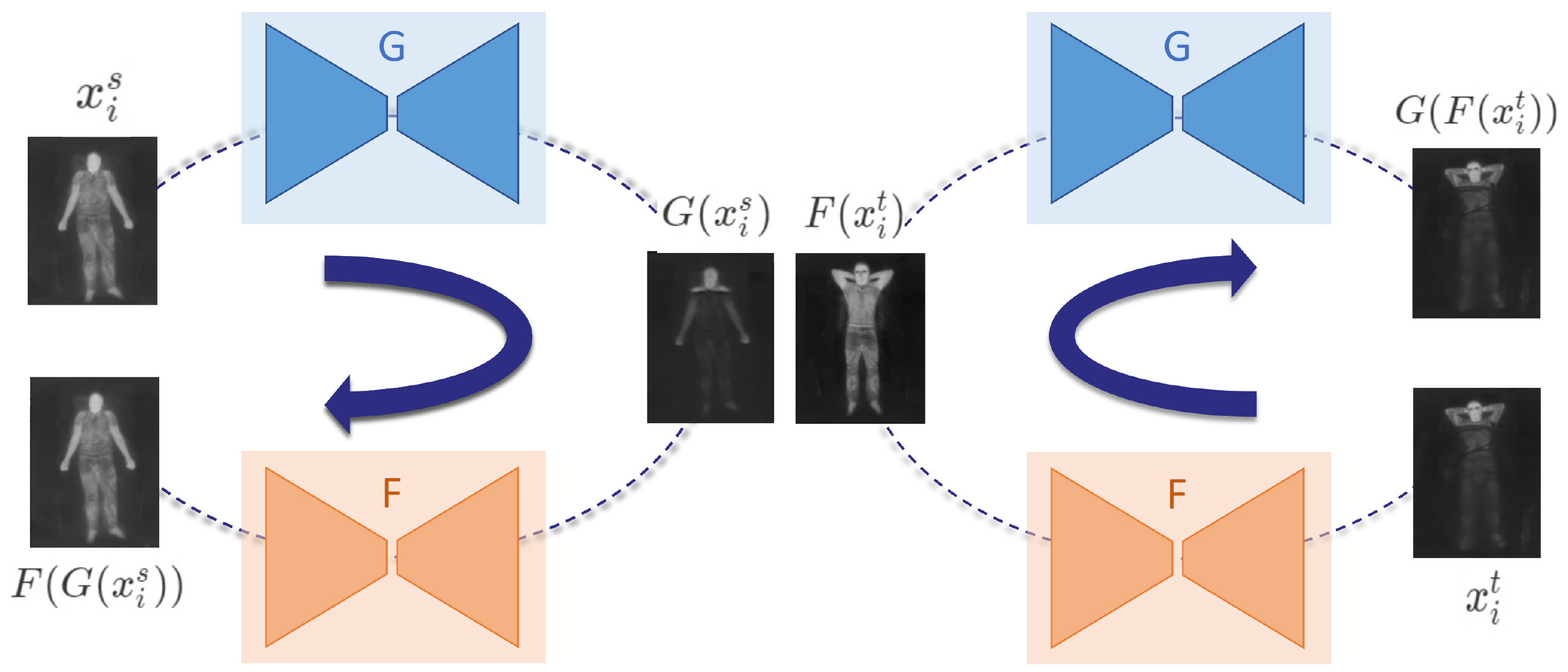}
    \vspace{-0.3em}
    \caption{CycAug: Translations from uncovered to covered and covered to uncovered are done separately from networks $G$ and $F$. Both networks are trained to learn the mappings in unsupervised manner.}
    \label{fig:cycaug}
    \vspace{-1em}
\end{figure}

\subsection{Two-fold Data Augmentation for Cross-Domain Discrepancy Reduction} \label{sec:data_aug}

To reduce the domain gap between $\mathcal{D}_s$ and $\mathcal{D}_{t_1}$, $\mathcal{D}_{t_2}$, we propose a data augmentation pipeline comprises of two key components; \textit{CycAug} and \textit{ExtremeAug}.

\vspace{0.4em}
\noindent \textbf{CycAug -- Unpaired Image-to-Image Translation based Data Augmentation:}
The problem becomes challenging when there are no labeled-covered LWIR images available to learn the mapping between poses and covered LWIR images. Here we propose \textit{CycAug}, based on the work done by Zhu \emph{et al.} \cite{Zhu2017UnpairedIT}, to convert a given image sampled from $\mathcal{D}_s$ to thin and thick covered LWIR images. The generated thin and thick covered images have minimal domain discrepancies with both the distributions $\mathcal{D}_{t_1}$ and $\mathcal{D}_{t_2}$, respectively. Through this, we are able to obtain a separate dataset that contains labeled thin covered LWIR images and labeled thick covered LWIR images. This dataset is further utilized to learn  a mapping between covered LWIR images and the underlying pose in a supervised manner with the additional domain discrepancy reduction techniques explained in the following sections.

The overview of the proposed \textit{CycAug} data augmentation is shown in Fig. \ref{fig:cycaug}. The objective of this is to find the mapping between $\mathcal{D}_s$ and $\mathcal{D}_t$ without having any data pairs. This can be formally defined as follows,
\begin{small}

\begin{equation}
\begin{aligned}
\mathcal{L}\left(G, F, D_{X}, D_{Y}\right) &=\mathcal{L}_{\mathrm{GAN}}\left(G, D_{Y}, X, Y\right) \\
&+\mathcal{L}_{\mathrm{GAN}}\left(F, D_{X}, Y, X\right) \\
&+\lambda \mathcal{L}_{\mathrm{cyc}}(G, F) +\lambda_{id} \mathcal{L}_{\mathrm{identity}}(G, F),
\end{aligned}
\end{equation}
where,

\vspace{-0.5em}

\begin{equation}
\begin{aligned}
\mathcal{L}_{\mathrm{GAN}}(G, D_{Y}, &X, Y) =\mathbb{E}_{y \sim Y}\left[\log D_{Y}(y)\right] \\
&+\mathbb{E}_{x \sim X}\left[\log \left(1-D_{Y}(G(x))\right]\right.
\end{aligned}
\end{equation}
\begin{equation}
\begin{aligned}
\mathcal{L}_{\mathrm{GAN}}(F, D_{X}, &Y, X) =\mathbb{E}_{x \sim X}\left[\log D_{X}(x)\right] \\
&+\mathbb{E}_{y \sim Y}\left[\log \left(1-D_{X}(F(y))\right]\right.
\end{aligned}
\end{equation}
\begin{equation}
\begin{aligned}
\mathcal{L}_{\mathrm{cyc}}(G, F) &=\mathbb{E}_{x \sim X}\left[\|F(G(x))-x\|_{1}\right] \\
&+\mathbb{E}_{y \sim Y}\left[\|G(F(y))-y\|_{1}\right]
\end{aligned}
\end{equation}
\begin{equation}
\begin{aligned}
\mathcal{L}_{\text {identity }}(G, F)&=\mathbb{E}_{y \sim Y}\left[\|G(y)-y\|_{1}\right]\\
&+\mathbb{E}_{x \sim X}\left[\|F(x)-x\|_{1}\right].
\end{aligned}
\end{equation}
%
%
\end{small}

Here $X, Y, G, F$ correspond to the uncovered images ($\{x_{i}^s\}_{i=1}^{n^s}$), covered images ($\{x_{i}^t\}_{i=1}^{n^t}$) and the generator networks utilized to do image to image translation from covered to uncovered and uncovered to covered, respectively. $D_X, D_Y$ correspond to discriminators utilized to make the $F$, $G$ better at generating realistic images. $\|.\|_{1}$ indicates the $\ell_1$ norm. Objective functions, $\mathcal{L}_{\mathrm{GAN}}(.)$, $\mathcal{L}_{\mathrm{cyc}}(.)$, $\mathcal{L}_{\mathrm{identity}}(.)$, $\mathcal{L}(.)$ represent the adversarial training of generator and discriminator to ensure the realistic image generations \cite{gan}, cycle consistency loss \cite{Zhu2017UnpairedIT} to learn the mapping between two image domains uncovered ($\{x_{i}^s\}_{i=1}^{n^s}$) and covered ($\{x_{i}^t\}_{i=1}^{n^t}$)  in an unsupervised manner, identity mapping regularization loss \cite{Zhu2017UnpairedIT} to regularize the learning and make convergence more efficient and the final loss as a weighted combination of above losses, respectively. For the training, we choose weighting factors $\lambda$ and $\lambda_{id}$ as $10$ and $5$, experimentally. 

After the training of networks $G$ and $F$, for both thin and thick cover LWIR images separately, $G_{\mathcal{D}_{t_1}}, F_{\mathcal{D}_{t_1}}, G_{\mathcal{D}_{t_2}}$ and $F_{\mathcal{D}_{t_2}}$ are obtained. Then we utilize $G_{\mathcal{D}_{t_1}}$ and $G_{\mathcal{D}_{t_2}}$ to generate thin and thick covered LWIR images separately given the uncovered images. The obtained generated covered images are further processed to reduce the domain discrepancies with $\mathcal{D}_{t}$.

\begin{figure}[t!]
    \centering
    \includegraphics[width = 0.92 \linewidth]{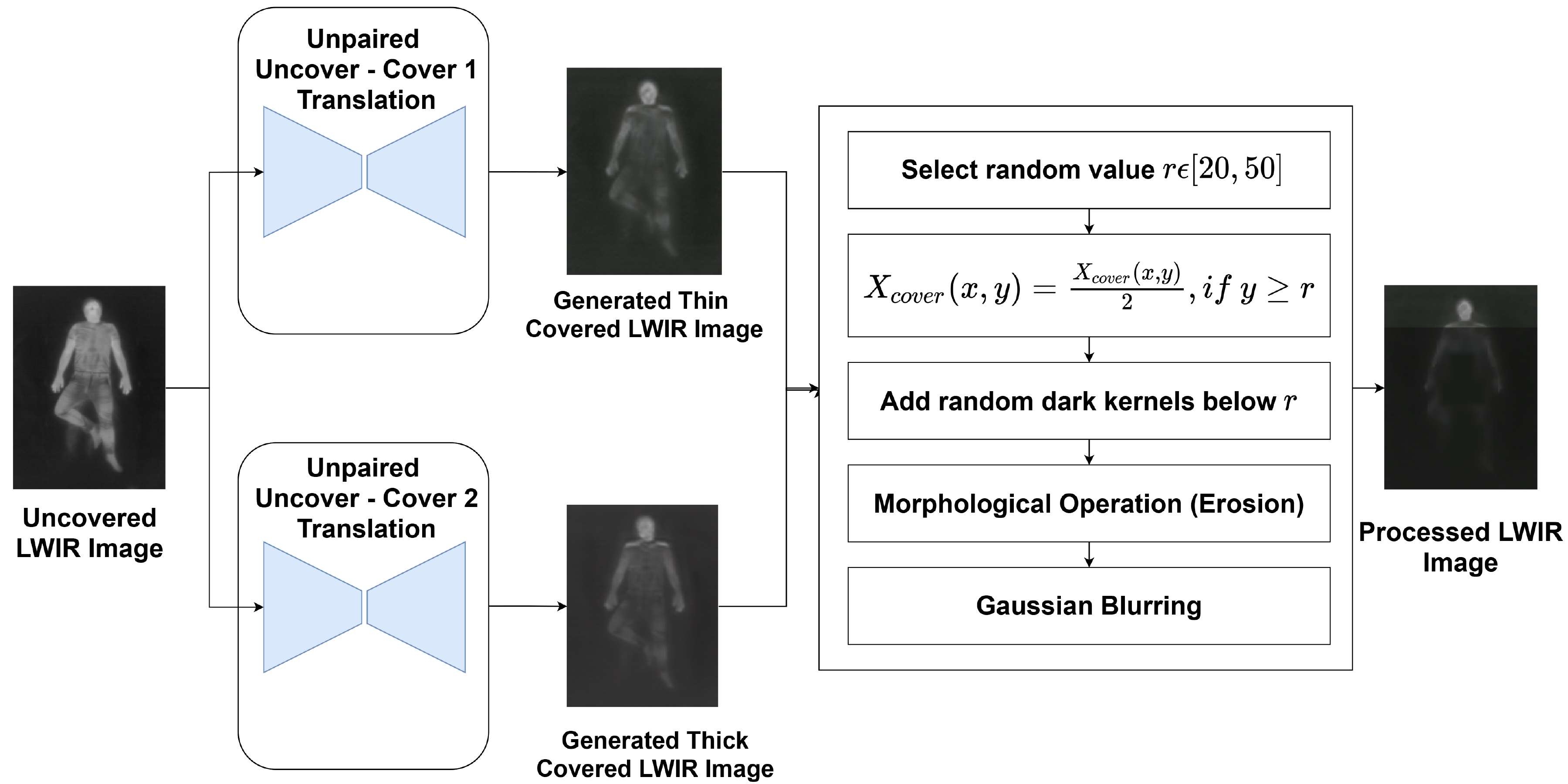}
    \caption{ExtremeAug: After obtaining the generated thin covered and thick covered images, a series of image processing techniques are applied to generate covered images with more occlusions.}
    \vspace{-1.5em}
    \label{fig:dataaug}
    \vspace{+0.4em}
\end{figure}


\vspace{0.4em}
\noindent \textbf{ExtremeAug -- Extreme Occlusion based Data Augmentation:}  Using \textit{ExtremeAug}, the generated thin and thick covered LWIR images are further augmented to have more covering artifacts / occlusions. By introducing different versions of covered images this way, we speculate that, the pose estimation model is expected to learn a robust mapping between covered images and pose, without becoming too sensitive to a certain single cover type (thick or thin). 


In \textit{ExtremeAug} as shown in Fig. \ref{fig:dataaug}, a random vertical point is selected in the second $\left(\frac{1}{8}\right)^{\text{th}}$ of the generated covered image from the top, and intensity of all the pixels below that value is decreased to mimic covered area of the body. The resultant image is further distorted by adding dark kernels (kernels with zero values) with the size of $20 \times 20$ on random areas of the image. Then the erosion morphological operation is applied using a kernel size of $3 \times 3$. Finally the image is blurred using Gaussian blurring to reduce visible information.
\vspace{-0.5em}

\subsection{Pose Estimation}   \label{sec:pose_estimation}



For an input image $x_{i}^s$ sampled from $\mathcal{D}_s$, we perform the two-fold data augmentation as described in Sec. \ref{sec:data_aug}. We denote the generated images from \textit{CycAug} as $x_{i}^{t_1}$ and $x_{i}^{t_2}$, where $t_1$ and $t_2$ corresponds to their respective domain distributions. The augmented image with \textit{ExtremeAug} are denoted as $x_{i}^{\Tilde{t}}$. We qualitatively validate (Fig. \ref{fig:dataaug}) that the domain discrepancy between distributions $\mathcal{D}_s$ and $\mathcal{D}_t$ are reduced by our two-fold data augmentation. Our intuition is that by reducing the domain discrepancy, the model tends to learn domain invariant features, which then facilitates in decoding covered domain features during inference. Further, by doing augmentations, we hypothesize the geometry of $x_{i}^s$, the augmented versions $x_{i}^{t_1}$, $x_{i}^{t_2}$ and $x_{i}^{\Tilde{t}}$ also have a ground truth label $y_{i}^s$. 

We extend the input space with augmented images and optimize the pose estimator $f_\theta$ using the standard mean squared error loss function $\mathcal{L}_{sup}$:
\begin{equation}
    \mathcal{L}_{sup}(f_\theta(x_i), y_i) = \frac{1}{\textit{K}}\sum_{j=1}^{\textit{K}} \vert\vert f_\theta(x_i) - y_i\vert\vert_{2}^{2}.
\end{equation}

\vspace{-0.5em}
\subsection{Knowledge Distillation} \label{sec:kd}
Knowledge Distillation \cite{kd} is a procedure of transferring the knowledge embedded in the teacher model to a student model. To this end, we take two clones of the best model trained previously and name them each as: $f_{\theta}^T$ and $f_{\theta}^S$, respectively. The weights of the teacher model, $f_{\theta}^T$ is frozen and used only during inference. $f_{\theta}^{T}(x_{i}^{t})$ is used as soft labels for the student model $f_{\theta}^S$. Through this, we enforce the student model to learn the distribution embedded with $\mathcal{D}_t$. In contrast to traditional knowledge distillation methods \cite{kd}, we employ standard mean squared error measure to enforce the knowledge distillation since our task is in a \emph{regression perspective}. Hence, the loss during this stage is given by:
\begin{equation}
    \mathcal{L}_{kd}(f_{\theta}^{S}(x_{i}^{t}), f_{\theta}^{T}(x_{i}^{t})) = \frac{1}{\textit{K}}\sum_{j=1}^{\textit{K}}  \vert\vert f_{\theta}^{S}(x_{i}^{t}) - f_{\theta}^{T}(x_{i}^{t})\vert\vert_2^{2}.
\end{equation}
We hypothesize that by learning the distilled knowledge from the teacher in a self-supervised manner, the student is expected to perform better than the teacher \cite{smooth}.


\section{Experiments and Results}
\label{sec:experiments}

\subsection{Dataset}    

We evaluate our proposed method on the dataset provided at the IEEE VIP Cup Challenge 2021\footnote[2]{\url{https://www.2021.ieeeicip.org/VIPCup.asp}}, which was extracted from the Simultaneously-collected multi-modal Lying Pose (SLP) dataset \cite{liu2020}. SLP dataset was collected from 109 participants under two different settings: 102 participants under home setting and 7 participants under hospital setting. During the experiment, the participants were asked to lie on the bed and allowed to randomly change their pose under three main sleep posture categories: left side, right side and supine. For each category, 15 poses were collected under 3 cover conditions (no cover, thin sheet $\sim$ 1mm and thick blanket $\sim$ 3mm) using 4 imaging modalities including RGB, depth, LWIR and pressure map, simultaneously.





In our study, training dataset consists of 1) uncovered and pose labeled LWIR images from $30$ subjects, 2) thin covered and pose unlabeled LWIR images from $25$ subjects and 3) thick covered and pose unlabeled LWIR images from $25$ subjects. There are no existing pairs of uncovered and covered images in the training data. Testing dataset consists of labeled thin and thick covered LWIR images from $10$ subjects.


\vspace{-0.5em}
\subsection{Implementation Details}     

For the implementation of our approach, we utilized stacked hourglass \cite{stackHG} and simple baseline \cite{simplebaseline} models as the baseline pose estimation networks. Stacked hourglass model consists of hourglass shaped networks stacked together. Each hourglass comprises of convolution, max pooling and upsampling layers to process the visual data and extract the important features. At the end of each hourglass network, two consecutive rounds of $1\times1$ convolutions are applied. Simple baseline model, on the other hand, comprises of an encoder which extracts visual features followed by a decoder to estimate poses. Following previous works \cite{stackHG, simplebaseline}, final network predictions are the heatmaps depicting the confidence of each joint in the given pixel coordinate. We use Adam optimizer with initial learning rate of $0.00025$, and the learning rate decays with a factor of $0.1$ in epochs $45$ and $60$. We train the standard supervision model for $100$ epochs while the knowledge distillation model is trained for $30$ epochs with a constant learning rate of $0.00025$.

\vspace{-0.5em}
\subsection{Results and Discussion}

\begin{table}[t]
\caption{PCKh@$0.5$ values on the test dataset. $^\dagger$ Results extracted from \cite{liu2020}.}
\centering
\begin{tabular}{p{5cm}|p{1.1cm}|p{1.2cm}}
    \hline
    \textbf{\multirow{2}{*}{\textbf{\hspace{2cm} Method}}} &\textit{\textbf{\small Newell et al. \cite{stackHG}}} & \textit{\textbf{\small Xiao \hspace{1cm} et al. \cite{simplebaseline}}} \\
    \hline
    Fully Supervised $^ \dagger$ & 94.0 & 92.5 \\
    Source (Uncover) data only & 39.52 & 24.12\\
    Uncover + CycAug & 72.44 & 70.43\\
    Uncover + CycAug + ExtremeAug & 73.38 & 71.87\\
    Uncover + CycAug + ExtremeAug + Knowledge Distillation & \textbf{76.13} & \textbf{73.84} \\
    \hline
    \end{tabular}
\label{tab:results}
\end{table}

The quantitative analysis on the different methods is shown in Table \ref{tab:results}. We also compare our performance with the fully-supervised method proposed in \cite{liu2020}. The experimental results clearly state the effectiveness of the proposed method in improving the performance by $36.61\%$ over the baseline in stacked hourglass model, showing the reduction of domain discrepancy between source and the target domains. The component-wise study of \textit{CycAug} and \textit{ExtremeAug} in Table \ref{tab:results} demonstrates that the heavy occlusion based data augmentation enhances the model to be robust to varying real world conditions. Similar improvement achieved in simple baseline \cite{simplebaseline} model ensures that our approach is \emph{agnostic to backbone architecture}.

Our proposed knowledge distillation improves the performance further by a margin of $2.75\%$ and $1.97\%$ in stacked hourglass and simple baseline respectively, in the PCKh@$0.5$ metric \cite{pck}. We attribute that the pseudo labels generated by the teacher model contribute to the performance enhancement of the student model. Since the knowledge distillation was performed in a self-supervised manner, it demonstrates that learning representations from unlabeled target domain is also crucial to boost the performance of the pose estimator. We believe that our approach motivates the extensive real-world data collection effort  instead of focusing on costly manual annotations.

\vspace{-1em}
\section{Conclusion}
\label{sec:conclusion}

In this paper, we present a domain adaption based training strategy to estimate in-bed human poses across varying illumination and occlusion conditions. Our results in two standard baseline pose estimation algorithms clearly indicate the reduction in cross-domain discrepancy and model agnostic characteristic of the method. This opens the door for future researches to invest effort on model agnostic pose estimation methods specific to in-bed setting. Looking forward, our work also opens up new directions for the largely unexplored and practically important cross-domain in-bed human pose estimation.

\vspace{-1em}
\section{Acknowledgements}
The authors extend their gratitude towards Dr. Ranga Rodrigo and the National Research Council of Sri Lanka for providing computational resources for conducting our experiments.

\bibliographystyle{IEEEtran}
\bibliography{main}

\end{document}